\documentclass[conference]{IEEEtran}
\IEEEoverridecommandlockouts
\usepackage{cite}
\usepackage{amsmath,amssymb,amsfonts}
\usepackage{graphicx}
\usepackage{textcomp}
\usepackage{xcolor}

\usepackage{mathrsfs}
\usepackage{algpseudocode, algorithmicx}
\usepackage{algorithm}
\usepackage{multicol}
\usepackage{multirow}
\usepackage{array}
\usepackage{booktabs}
\usepackage{float}
\usepackage{subfig}
\usepackage{colortbl}
\usepackage{makecell}

\def\BibTeX{{\rm B\kern-.05em{\sc i\kern-.025em b}\kern-.08em
    T\kern-.1667em\lower.7ex\hbox{E}\kern-.125emX}}
\begin{document}

\title{MADLLM: Multivariate Anomaly Detection  via Pre-trained LLMs}

\author{\IEEEauthorblockN{
    Wei Tao$^{\spadesuit}$$^{\heartsuit}$,
    Xiaoyang Qu$^{\heartsuit*}$,
    Kai Lu$^{\spadesuit*}$\thanks{$^{*}$Xiaoyang Qu (email: quxiaoy@gmail.com) and Kai Lu (email: kailu@hust.edu.cn) are the corresponding authors.},
    Jiguang Wan$^{\spadesuit}$,
    Guokuan Li$^{\spadesuit}$,
    Jianzong Wang$^{\heartsuit}$}
    \IEEEauthorblockA{$^{\spadesuit}$Huazhong University of Science and Technology, Wuhan, China}
    \IEEEauthorblockA{$^{\heartsuit}$Ping An Technology (Shenzhen) Co., Ltd., Shenzhen, China}
}

\maketitle

\begin{abstract}
When applying pre-trained large language models (LLMs) to address anomaly detection tasks, the multivariate time series (MTS) modality of anomaly detection does not align with the text modality of LLMs. Existing methods simply transform the MTS data into multiple univariate time series sequences, which can cause many problems. This paper introduces MADLLM, a novel multivariate anomaly detection method via pre-trained LLMs. We design a new triple encoding technique to align the MTS modality with the text modality of LLMs. 
Specifically, this technique integrates the traditional patch embedding method with two novel embedding approaches: ({\romannumeral1}) Skip Embedding, which alters the order of patch processing in traditional methods to help LLMs retain knowledge of previous features, and ({\romannumeral2}) Feature Embedding, which leverages contrastive learning to allow the model to better understand the correlations between different features. Experimental results demonstrate that our method outperforms state-of-the-art methods in various public anomaly detection datasets.
\end{abstract}


\section{Introduction}
\label{sec:intro}

Anomaly detection is widely used across a range of fields, including finance. Previous scholars have relied on classic machine learning methods, deep neural network architectures, and attention-based models to handle anomaly detection tasks, yielding remarkable breakthroughs. However, due to concerns about business secrets, enterprises are often unwilling to disclose their anomaly data. This causes a severe shortage of labeled anomaly data, which poses a significant challenge for anomaly detection tasks. In this case, traditional models usually face unknown pattern anomalies and require a complex redesign and lengthy retraining. Recently, some researchers have proposed to leverage pre-trained large language models (LLMs) to detect time series anomalies, using their inherent abstraction and generalization capabilities \cite{zhou2023one, wang2024incprompt}. The unified framework and strong generalization capabilities provided by the pre-trained LLMs effectively solve the problem of the lack of labeled data prevalent in anomaly detection tasks.
\begin{figure}[t]
    \centering
    \includegraphics[width=0.48\textwidth]{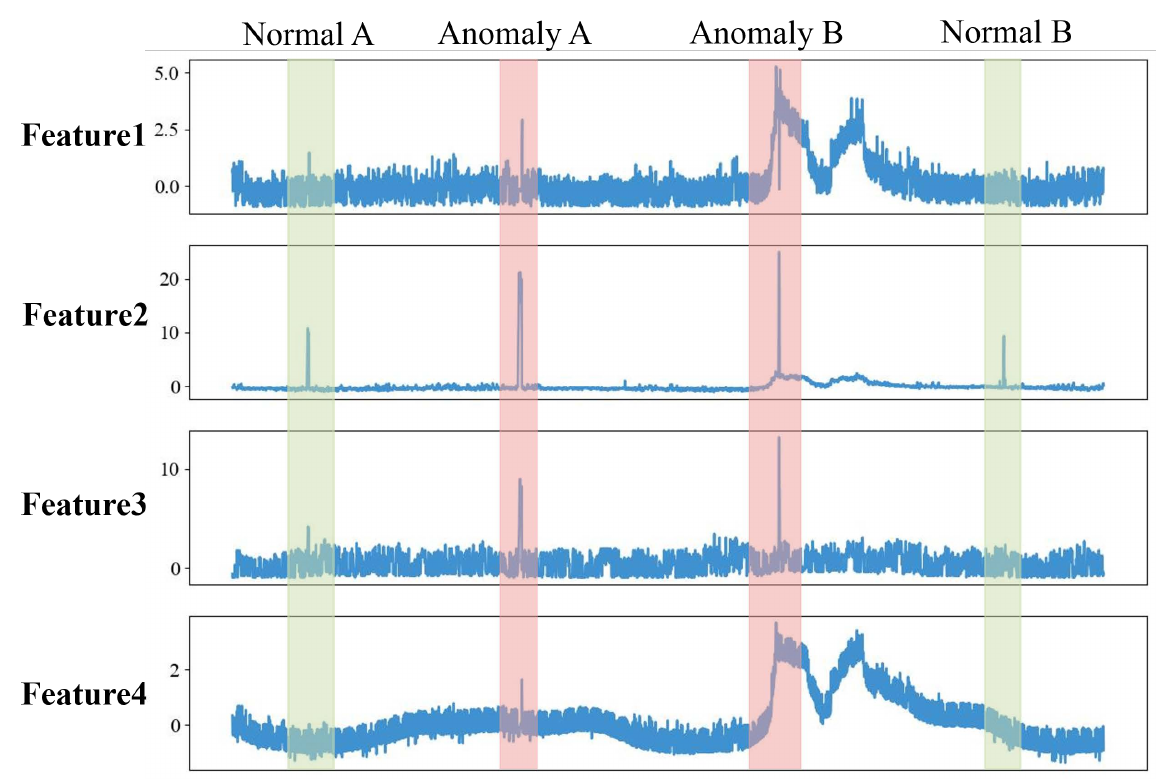}
    \caption{An example of anomalies in the SMD dataset. The four features are marked as \textbf{Feature1-4} due to dataset privacy.}
    \label{fig:challenge}
    \vspace{-6mm}
\end{figure}

The problem of anomaly detection via pre-trained LLMs lies in the encoding method of the input data. The original input modality of LLMs is discrete text, while the input modality of the anomaly detection task is continuous multivariate time series (MTS) data. The two modalities cannot be aligned, which can reduce the accuracy of the model. To solve this problem, researchers have proposed a patch embedding method to encode the MTS data\cite{zhou2023one}. The current patch embedding method treats the multivariate time series data as multiple univariate time series sequences, concatenates them serially, and inputs them into the pre-trained LLMs. However, this approach has two challenges. First, it ignores the correlations among different features (we call each variate in the MTS data a feature), which can lead to misjudgment. Anomalies are often accompanied by problems with multiple features, and anomaly detection requires combining the observations of multiple features to make judgments. For example, we extract a segmentation from a famous MTS dataset SMD \cite{su2019robust} and illustrate it in Figure \ref{fig:challenge}. In the time period marked as Normal A, \textbf{Feature1-3} all have a spike, which can be mistakenly classified as an anomaly if we do not pay attention to \textbf{Feature4}. A similar
situation happens to Normal B, where only \textbf{Feature2} has a spike. We need to consider all the features to detect a true anomaly, such as Anomaly A and Anomaly B. Second, the input sequence becomes very long after the serial concatenation of multiple feature sequences. Therefore, when the model processes the feature data at the end of the input sequence, it may forget the feature information at the beginning of the input sequence.

To address the challenges described above, in this paper, we propose a new method for \underline{m}ultivariate \underline{a}nomaly \underline{d}etection problems based on pre-trained \underline{LLM}s, i.e., MADLLM. We design a new triple encoding technique to encode the MTS input data, which can help the LLMs better maintain the memory of distant historical feature information and understand the correlations among various features. Specifically, the triple encoding technique incorporates the patch embedding method in previous works with two novel embedding approaches: \textbf{skip embedding} and \textbf{feature embedding}. Skip embedding is achieved by rearranging the patch processing order in traditional patch embedding. We skip to patches of other features every time after processing the patch of one feature so that the model does not forget information about other features just because they are too far away on the input sequence. Feature embedding is achieved by a contrastive learning encoder. We use a causal convolutional network as the network backbone and define a patch-based triplet loss to update the encoder. 

The contributions of our work are summarized as follows:
\begin{itemize}
    \item We design a new triple encoding technique for more accurate anomaly detection with MTS data.
    \item We propose skip embedding to prevent LLMs from forgetting features from distant past inputs.
    \item We propose feature embedding to capture the correlations among different features.
    \item Extensive experiments conducted on several public datasets illustrate that our method outperforms state-of-the-art anomaly detection methods.
\end{itemize}
\section{Related Works}
\label{sec:related works}

\subsection{Anomaly Detection via Traditional Models}
Anomaly detection has always been a widespread problem in both industrial and research areas. Traditional models for anomaly detection can be summarized as three kinds. First, some researchers propose to use traditional machine learning methods to tackle anomaly detection. MERLIN \cite{nakamura2020merlin} employs a parameter-free approach to detect time series anomaly by iteratively comparing subsequences of different lengths with their adjacent neighbors. Second, some researchers address anomaly detection problems via deep neural networks \cite{zong2018deep, su2019robust, audibert2020usad,  wutimesnet}, including convolutional neural networks (CNN), recurrent neural networks (RNN) and long-short term memory (LSTM). For example, OmniAnomaly \cite{zong2018deep} reconstructs input data by the representations of the normal pattern of MTS and uses the reconstruction probabilities to determine anomalies. Third, models based on transformer modules or attention mechanisms have emerged recently \cite{ li2021opengauss, chen2023imdiffusion, tuli2022tranad}. For instance, Tuli et al. \cite{tuli2022tranad} propose TranAD, a deep transformer network-based anomaly detection and diagnosis model that uses attention-based sequence encoders to swiftly perform inference with the knowledge of the broader temporal trends in the data. However, these models require complex redesign and lengthy retraining periods when facing unprecedented anomaly types. In this paper, we propose to use pre-trained LLMs to address anomaly detection problems. With the strong generalization capabilities of the pre-trained LLMs, we only need to fine-tune the model in the case of novel anomaly patterns. 

\subsection{LLMs for Time Series Data Tasks}
Recent years have witnessed the emergence of a series of LLMs \cite{cocktail}. Therefore, researchers have proposed to apply LLMs to the field of anomaly detection \cite{zhou2023one, sun2023test, liu2024large, liu2024large2}. For instance, 
Zhou et al. \cite{zhou2023one} propose to input the MTS data into a pre-trained GPT2 model and fine-tune the model to achieve anomaly detection results. AnomalyLLM \cite{liu2024large} distills the LLM into a student network. During inference, when the features of the student network are greatly different from those of the LLM, anomalies are indicated. LLMAD \cite{liu2024large2} leverages positive and negative similar time series segments and an Anomaly Detection Chain-of-Thought approach to accurately predict anomalies. However, these methods treat MTS as multiple univariate series, which makes the model forget distant history feature information and neglect the correlations among features. This paper proposes two optimizations (skip embedding and feature embedding) to avoid these problems.

\section{Proposed Method}
\label{sec: proposed}
In this section, we first introduce the overall architecture of MADLLM. Then, we describe the two important optimizations in MADLLM: skip embedding and feature embedding.
\subsection{Architecture}
\begin{figure}[t]
    \centering
    \includegraphics[width=0.48\textwidth]{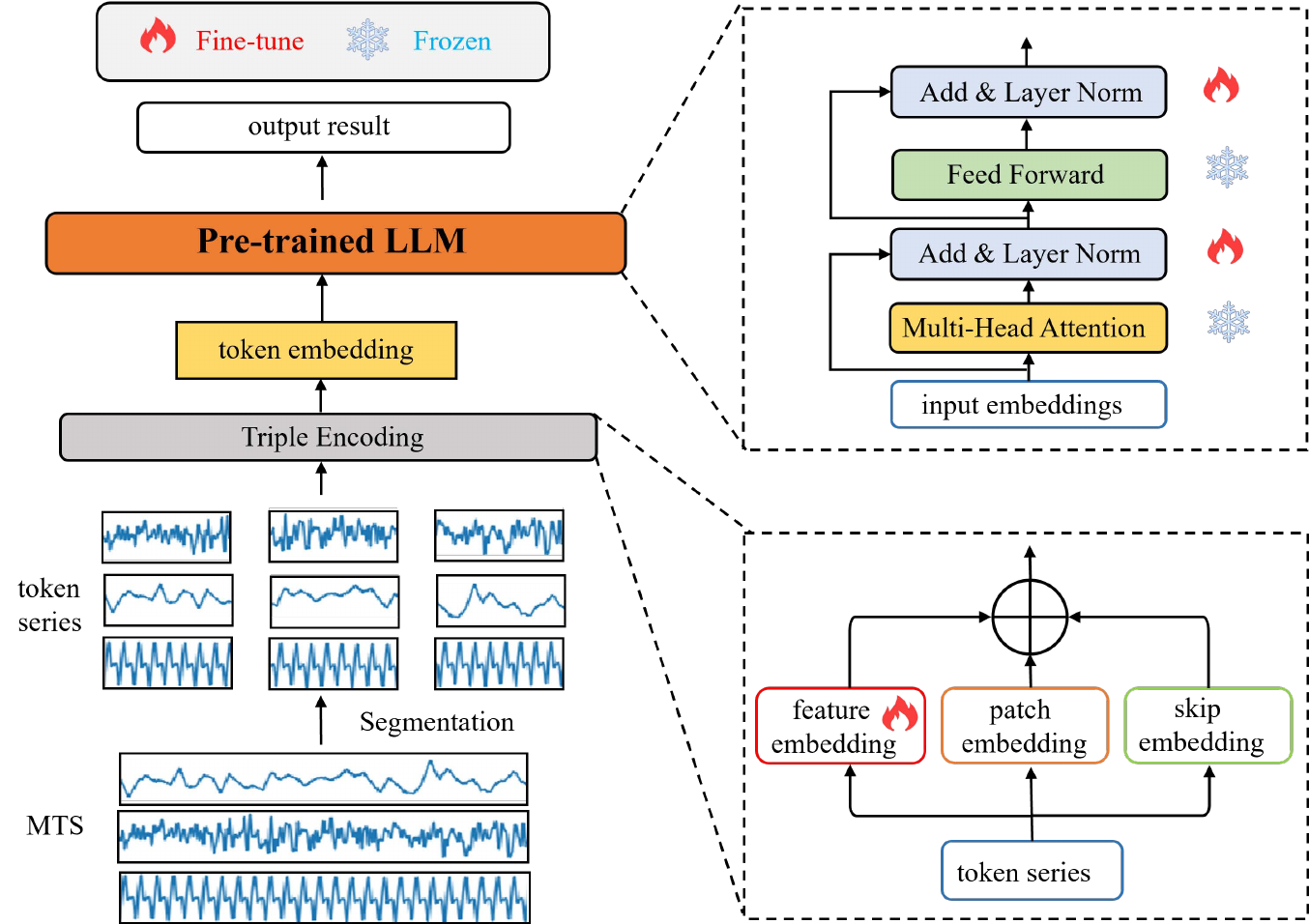}
    \caption{The architecture of MADLLM.}
    \label{fig:framework}
    \vspace{-5mm}
\end{figure}
Figure \ref{fig:framework} shows the architecture of our MADLLM method. The MTS data is first segmented into a token series according to the temporal order, where the data of each feature is divided into equal-sized patches. Then, we use triple encoding to encode the token series into three embeddings, including traditional patch embedding, skip embedding, and feature embedding. We integrate the three embeddings to form the final input of the pre-trained LLM, which is called token embedding.
During the inference process of MADLLM, we freeze the multi-head attention and feed forward layers in the pre-trained LLM body while periodically fine-tuning the normalization layers and the feature embedding module.
This is because the multi-head attention and feed forward layers contain the majority of learned knowledge from pre-trained LLMs. Freezing them can not only maintain the representation ability of the model but also save lots of fine-tuning time.

\subsection{Skip Embedding}
\begin{figure}[t]
    \centering
    \includegraphics[width=0.48\textwidth]{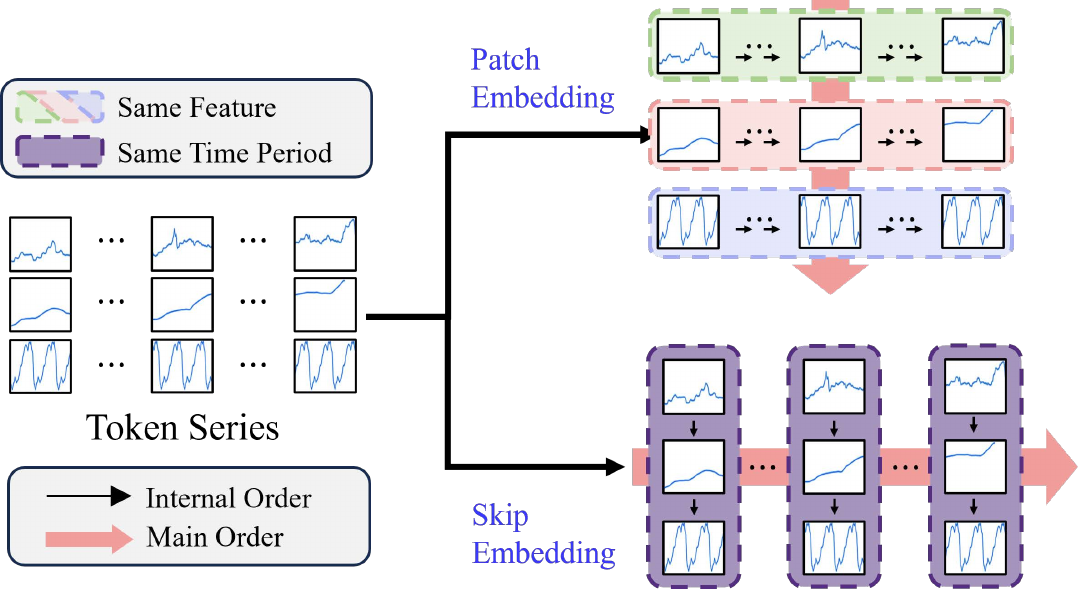}
    \caption{
    The workflow of skip embedding and traditional patch embedding. 
    }
    \label{fig:skip embedding}
    \vspace{-5mm}
\end{figure}

Figure \ref{fig:skip embedding} illustrates a demonstration workflow of traditional patch embedding and skip embedding. For the token series described above, the main order of traditional   
patch embedding is from feature to feature. It converts the sequence of patches within each feature into embeddings, and after the conversion of data within one feature is completed, it proceeds to the following feature for conversion. In contrast, the main order of our skip embedding approach is from time period to time period. After converting a patch within a specific feature into an embedding, we ``skip'' to the next feature within the same time period to continue conversion.
For example, first, the model selects a patch from the series of the first feature, then it skips to the series of the second feature and selects a patch in the same time period, then the third feature. 
This is why our method is called \textbf{skip embedding}. After processing patches of all the features in this time period, it turns to the following time period and continues the loop.  

During inference, to reduce the latency in the ``skip" process, we reorder the patches in the original token series to ensure that the patches in the same time period are contiguous in physical memory. Suppose that the original token series is:
\begin{equation}
    \mathcal{S} = \{\mathbf{s_1^1}, \mathbf{s_2^1},\cdots, \mathbf{s_P^1}, \mathbf{s_1^2},\cdots,\mathbf{s_P^2},\cdots,\mathbf{s_P^M}\}
\end{equation}
where $s_i^j$ is the $i_{th}$ patch in the $j_{th}$ feature, $P$ and $M$ are the number of patches and features, respectively. 
Then, the new token series is as follows:
\begin{equation}
    \mathcal{E} = \{\mathbf{s_1^1}, \mathbf{s_1^2},\cdots, \mathbf{s_1^M}, \mathbf{s_2^1},\cdots, \mathbf{s_2^M},\cdots, \mathbf{s_P^M}\}
\end{equation}

After reordering the patches in the token series, we convert the new token series $\mathcal{E}$ with the same embedding function as how patch embedding is achieved, and we finally get the skip embedding. Patch embedding captures the information of a single feature over different time periods, while skip embedding captures the information of different features at a single time period. Skip embedding helps prevent the model from forgetting information from features that are in the distant past, while patch embedding compensates for the difficulty of skip embedding in capturing cross-time period information. These two types of embeddings are complementary. Therefore, we combine them to achieve better anomaly detection results.

\subsection{Feature Embedding}
\label{subsec: feature embedding}
Since contrastive learning does not rely on prior knowledge, it will not be influenced by the lack of labeled anomaly data \cite{sun2023test}. Therefore, we employ contrastive learning to learn the correlations among different features and achieve a new embedding called feature embedding. We introduce an additional deep learning network called contrastive learning encoder (independent from the pre-trained LLM) to generate the feature embedding. The encoder is randomly initialized and will be updated at the fine-tuning stage of MADLLM.

To learn the correlations among different features, we need to make the representations of the data from similar features closer while increasing the distance among the representations of data from dissimilar features. Inspired by the idea of \cite{franceschi2019unsupervised}, we propose a patch-based triplet loss to update our contrastive learning encoder. We denote patches from the same feature as similar, while patches from random different features are denoted as dissimilar. 

\begin{algorithm}[t]
    \caption{The calculation of our patch-based triplet loss in contrastive learning in a fine-tuning epoch.}
    \label{alg1}
    \begin{algorithmic}[1]
    \Require Token series $\mathcal{S}$, Number of features $M$
    \Ensure Patch-based triplet loss of the contrastive learning $L$
    \For{$i = 1$ to $M$}
    \State Denote the number of patches in a feature as P
        \State Randomly choose $s_p^i$ as the $s_{anc}$, $1 \leq p \leq P$
        \State Randomly choose another $s_q^i$ as the $s^+$, $1 \leq q \leq P$, $p \neq q$, 
        \State Randomly choose N number from $[1, M]$, $i_j \neq i, j=0,1,...,N$
        \For{$j = 1$ to $N$}
            \State Randomly choose $s_p^{i_j}$ as the $s_j^-$, $1 \leq p \leq P$
        \EndFor
        \State Calculate triplet loss $L_i$ by equation 7
    \EndFor
    \State $L \leftarrow$ Concatenate all the $L_i$ 
    \State return $L$
    \end{algorithmic}
    
\end{algorithm}

Algorithm \ref{alg1} demonstrates how we calculate the patch-based triplet loss in contrastive learning. For each feature, we randomly sample a patch from the feature and denote it as the anchor patch, named $s_{anc}$. Then, we randomly choose another patch from the same feature as the positive patch, named $s^+$. Furthermore, we randomly sample $N$ features ($N$ is a pre-defined hyperparameter) from the rest of the features and randomly select one patch from each feature as a negative patch, named $\{s_j^-\}_{j=1}^N$. Given the effectiveness and efficiency of InfoNCE \cite{he2020momentum} in minimizing the difference among positive pairs and maximizing the difference among negative pairs in the embedding space, we adopt the InfoNCE algorithm to calculate our triplet loss. The patch-based triplet loss for one feature is calculated as follows:

\begin{equation}
    L = -log\frac{exp(f(s_{anc}, s^+))}{exp(f(s_{anc}, s^+)) + \sum_{j=1}^{N}{exp(f(s_{anc}, s_j^-))}}
\end{equation}
where $f()$ is the cosine similar function. Specifically, $f()$ is:
\begin{equation}
    f(u,v) = \frac{u^Tv}{||u||_2\cdot||v||_2}
    \label{eqn:cosine}
\end{equation}
where $||u||_2$ means the L2 norm of vector $u$. We travel through all the features, find such $(s_{anc}, s^+, s^-)$ triple tuple, and calculate patch-based triplet loss for each feature. Finally, we concatenate the patch-based triplet loss of all the features to form the patch-based triplet loss of the whole token series, which will be used to update our contrastive learning encoder.

After we achieve the patch-based triplet loss of the whole token series, we need to determine a network backbone to execute the contrastive learning task and achieve the feature embedding. Exponentially causal convolutional networks \cite{bai2018empirical} are neural network architectures that combine causal convolutions and exponential decay mechanisms. They avoid the vanishing and exploding gradient problems associated with RNNs while ensuring the causality of time series data, a property that ordinary CNNs lack. This makes them particularly well-suited for handling time series data. Therefore, we choose the exponentially causal convolution network as our contrastive learning encoder backbone. 
\begin{figure}[t]
    \centering
    \includegraphics[width=0.48\textwidth]{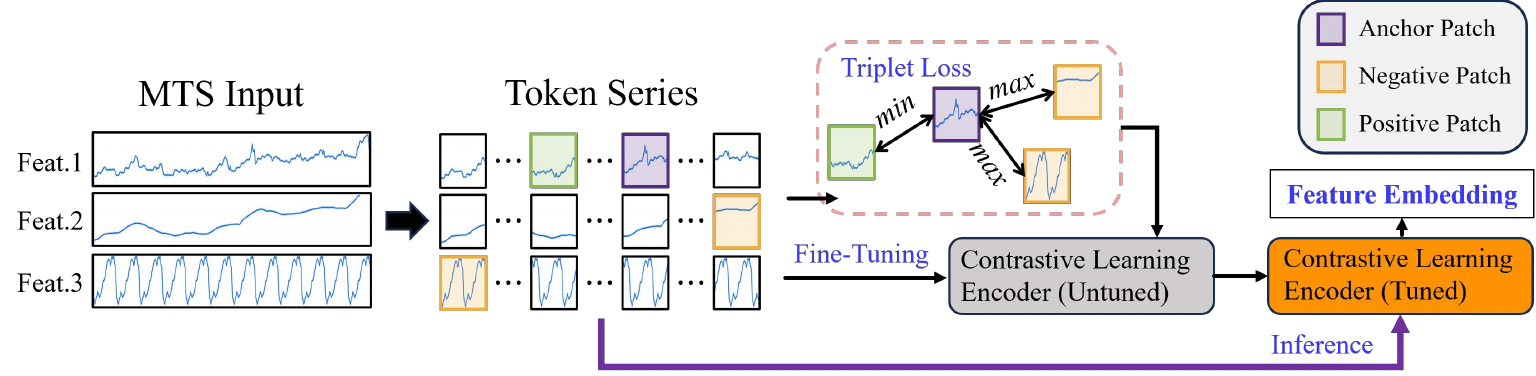}
    \caption{The workflow of feature embedding.}
    \label{fig:feature embedding}
    \vspace{-7mm}
    
\end{figure}

Figure \ref{fig:feature embedding} demonstrates the workflow of feature embedding. During the fine-tuning stage of MADLLM, we use the token series as the input and the patch-based triplet loss as the loss to update the contrastive learning encoder. The fine-tuning of the contrastive learning encoder is independent of the fine-tuning of the LLM body. Since the network of the contrastive learning encoder does not have too many layers, the fine-tuning process is lightweight. During the inference stage, we input the token series into the updated contrastive learning encoder and achieve the feature embedding.
\section{Experiments}
\label{sec: evaluation}
In this section, we conduct a comprehensive evaluation of MADLLM. 

\subsection{Experimental Setup}

\textbf{Datasets.} We compare our method with the baseline methods on five widely used public time-series anomaly detection datasets, all of which are multivariate. The datasets are: (1) SMD \cite{su2019robust}, which is collected from 28 different machines from a large Internet
company. (2) PSM \cite{abdulaal2021practical}, which is collected internally from
multiple application server nodes at eBay. (3) SWaT \cite{mathur2016swat}, which is collected from a real-world water treatment plant. (4) SMAP \cite{hundman2018detecting}, which is a dataset of soil samples and telemetry information labeled
by the SMAP satellite of NASA. (5) MSL \cite{hundman2018detecting}, which is similar to SMAP, but it is labeled by the MSL rover of NASA. To ensure the fairness of the experiments, we test the models on all the subsets of these datasets rather than only on some non-trivial subsets \cite{tuli2022tranad}, so the results of some baseline methods may be different from the statistics in their paper. The details of the datasets are shown in Table \ref{tab:dataset}.

\begin{table}[t]
    \centering
    \caption{Information of Evaluation Datasets.}
    \begin{tabular}{lcccc}
    \toprule
        Dataset & Train & Test & Features & Anomalies (\%) \\
        \midrule
        SMD\cite{su2019robust} & 708405 & 708420 & 38 & 4.16 \\
        PSM\cite{abdulaal2021practical}& 132481 & 87841 & 25 & 1.0 \\
        SWaT\cite{mathur2016swat} & 496800 & 449919 & 51 & 11.98 \\
        SMAP\cite{hundman2018detecting} & 153183 & 427617 & 25 & 13.13 \\
        MSL\cite{hundman2018detecting} & 58317 & 73729 & 55 & 10.72 \\
        \bottomrule
        
    \end{tabular}
    \vspace{-5mm}
    \label{tab:dataset}
    
\end{table}

\textbf{Baselines.} We compare our method with several current state-of-the-art anomaly detection solutions that use different models, including classic machine learning methods (MERLIN \cite{nakamura2020merlin}), traditional deep neural networks (OmniAnomaly \cite{su2019robust}, USAD \cite{audibert2020usad}, TimesNet \cite{wutimesnet}), attention mechanism-based models (
TranAD \cite{tuli2022tranad}, ImDifussion \cite{chen2023imdiffusion}), and methods based on pre-trained LLMs (SFT \cite{zhou2023one}, AnomalyLLM \cite{liu2024large}, LLMAD \cite{liu2024large2}).

\textbf{Evaluation Metrics.} Anomaly detection is a binary classification task, and we need to judge if the time period is normal or anomalous. Therefore, we use \textbf{F1 score} and \textbf{AUC} as the evaluation metrics.

\begin{table*}[ht]
\centering

\caption{The performance comparison of different anomaly detection methods on different datasets.}
\label{table:performance}
  \begin{tabular}{l@{\hspace{0.5cm}}cc@{\hspace{0.5cm}}cc@{\hspace{0.5cm}}cc@{\hspace{0.5cm}}cc@{\hspace{0.5cm}}cc@{\hspace{0.5cm}}}
    \toprule[1pt]
    \specialrule{0em}{1pt}{1pt}
    \multirow{2}{*}{Method}  & \multicolumn{2}{c}{SMD} & \multicolumn{2}{c}{PSM} & \multicolumn{2}{c}{SWaT} & \multicolumn{2}{c}{SMAP} & \multicolumn{2}{c}{MSL} \\
    \specialrule{0em}{1pt}{1pt}
    \cline{2-11}
    \specialrule{0em}{1pt}{1pt}
       & F1 $\uparrow$ & AUC  $\uparrow$ & F1 $\uparrow$ & AUC $\uparrow$ & F1 $\uparrow$ & AUC $\uparrow$ & F1 $\uparrow$ & AUC $\uparrow$ & F1 $\uparrow$ & AUC $\uparrow$\\
    \specialrule{0em}{1pt}{1pt}
    \hline
    \specialrule{0em}{1pt}{1pt}

    MERLIN\cite{nakamura2020merlin} & 0.3842 & 0.7158 & 0.4702 & 0.7245 & 0.3669 & 0.6175 & 0.2724 & 0.7426 & 0.3345 & 0.6281 \\

    OmniAnomaly\cite{su2019robust} & 0.8899 & 0.9401  & 0.9192 & 0.9045  & 0.8061 & 0.8467 & 0.8996 & 0.9051 & 0.8222 & 0.8875\\

    USAD\cite{audibert2020usad}  & 0.8991 & 0.9495 & 0.9146 & 0.9477  & 0.7789 & 0.8460 & 0.7703 & 0.8752 & 0.8509 & 0.9057\\

    TimesNet\cite{wutimesnet} & 0.8587 & 0.9032  & 0.9602 & 0.9455  & 0.8824 & 0.8973 & 0.7152 & 0.8435 & 0.8517 & 0.8246\\
    
    TranAD\cite{tuli2022tranad} & 0.8944 & 0.9362  & 0.9220 & 0.9457 & 0.7143 & 0.8491 & 0.8361 & 0.8925 & 0.9121 & 0.9344\\ 
    
    ImDifussion\cite{chen2023imdiffusion}& \textbf{0.9488} & 0.9346 & 0.9782 & 0.9684  & 0.8719 & 0.9233 & 0.9175 & 0.9248 & 0.8782 & 0.9091\\
    
    SFT\cite{zhou2023one} & 0.8388 & 0.9012 & 0.9686 & 0.9724  & 0.8887 & 0.9263 & 0.6888 & 0.7743 & 0.8391 & 0.9133\\

    AnomalyLLM \cite{liu2024large} & 0.9332 & 0.9568 & 0.9714 & 0.9728 & 0.8875 & 0.9248 & 0.9149 & 0.9237 & 0.9103 & 0.9211\\

    LLMAD \cite{liu2024large2} & 0.9142 & 0.9535 & 0.9633 & 0.9689 & 0.8815 & 0.9243 & 0.9177 & 0.9238 & 0.9088 & 0.9226 \\
    
    \rowcolor{orange!40}
    MADLLM & 0.9372 & \textbf{0.9872} & \textbf{0.9787} & \textbf{0.9852}  & \textbf{0.9138} & \textbf{0.9451} & \textbf{0.9323} & \textbf{0.9668} & \textbf{0.9237} & \textbf{0.9838}\\

    \bottomrule[1pt]
  \end{tabular}

\end{table*}

\begin{table*}[ht]
    \centering
    \caption{The performance comparison of different anomaly detection methods using \underline{20\%} training data.}
    \begin{tabular}{l@{\hspace{0.5cm}}cc@{\hspace{0.5cm}}cc@{\hspace{0.5cm}}cc@{\hspace{0.5cm}}cc@{\hspace{0.5cm}}cc@{\hspace{0.5cm}}}
    \toprule[1pt]
        \multirow{2}{*}{\textbf{Method}} &\multicolumn{2}{c}{SMD} & \multicolumn{2}{c}{PSM} & \multicolumn{2}{c}{SWaT} & \multicolumn{2}{c}{SMAP} & \multicolumn{2}{c}{MSL}\\
    \cmidrule{2-11}

    & F1 $\uparrow$ & AUC $\uparrow$ & F1 $\uparrow$ & AUC $\uparrow$ & F1 $\uparrow$ & AUC $\uparrow$ & F1 $\uparrow$ & AUC $\uparrow$ & F1 $\uparrow$ & AUC $\uparrow$ \\
    
    \midrule
        MERLIN \cite{nakamura2020merlin} & 0.3202 & 0.6358 & 0.4091 & 0.6286 & 0.3019 & 0.5284 & 0.2346 & 0.6989 & 0.2675 & 0.5542\\ 
        OmniAnomaly \cite{su2019robust}& 0.8064 & 0.7024 & 0.8648 & 0.8558 & 0.7433 & 0.7619 & 0.8131 & 0.8079 & 0.7409 & 0.7863\\ 
        USAD \cite{audibert2020usad}& 0.8234 & 0.8056 & 0.8664 & 0.8257 & 0.7087 & 0.7938 & 0.7328 & 0.7245 & 0.5198 & 0.8413\\
        
        TimesNet\cite{wutimesnet} & 0.6625 & 0.8134 & 0.5718 & 0.8649 & 0.4145 & 0.6690 & 0.5418 & 0.7007 & 0.5718 & 0.7812\\ 
        TranAD \cite{tuli2022tranad}& 0.8407 & 0.8894 & 0.8603 & 0.8566 & 0.6594 & 0.7738 & 0.7728 & 0.7878 & 0.8380 & 0.8652 \\ 
        Imdifussion\cite{chen2023imdiffusion} & 0.7816 & 0.8049 & 0.7918 & 0.8031 & 0.8001 & 0.8421 & 0.8369 & 0.7563 & 0.6018 & 0.5728\\ 
        SFT\cite{zhou2023one} & 0.8027 & 0.8596 & 0.9236 & 0.9317 & 0.8850 & 0.9204 & 0.6675 & 0.7538 & 0.7547 & 0.8687 \\ 

        AnomalyLLM \cite{liu2024large} & 0.9204 & 0.9433 & 0.9135 & 0.9224 & 0.8735 & 0.9123 & 0.9127 & 0.9112 & 0.9090 & 0.9041\\

    LLMAD \cite{liu2024large2} & 0.9086 & 0.9488 & 0.9127 & 0.9215 & 0.8744 & 0.9213 & 0.9062 & 0.9154 & 0.9069 & 0.8937\\
    
        \rowcolor{orange!40}
        MADLLM & \textbf{0.9265} & \textbf{0.9782} & \textbf{0.9268} & \textbf{0.9381} & \textbf{0.9098} & \textbf{0.9447} & \textbf{0.9302} & \textbf{0.9663} & \textbf{0.9203} & \textbf{0.9547}\\ 

    \bottomrule[1pt]
    \end{tabular}
    \label{table:few shot}
    \vspace{-2mm}
\end{table*}
\begin{table*}[t]
    \centering
    \caption{The training time comparison of different anomaly detection methods on SMD dataset.}
    \vspace{-2mm}
    \begin{tabular}{ccccccccccc}
    \toprule[1pt]
    Methods & MERLIN & OmniAnomaly & USAD & TimesNet & TranAD & ImDifussion & SFT & AnomalyLLM & LLMAD & \cellcolor{orange!40}MADLLM \\
    \midrule
    Training Time (s) $\downarrow$  & 72.32 & 276.97 & 250.97 & 346.86 & 43.56 & 303.49 & \textbf{1.32} & 160.38 & 86.45 & \cellcolor{orange!40}1.89  \\
    \bottomrule[1pt]
    \end{tabular}
    \vspace{-5mm}
    \label{tab:training time}
\end{table*}
\subsection{Evaluation Results}

We compare MADLLM with the baseline models on all the five datasets discussed above. As is illustrated in Table \ref{table:performance}, the average F1 score and AUC of our MADLLM model are 0.9371 and 0.9736, respectively. Our MADLLM model outperforms all the baseline methods over all five datasets except SMD on the F1 score evaluation metric. This might be because the distribution bias between normal and anomalous data in this dataset is relatively small, and ImDiffusion's unconditional attribution algorithm is able to amplify the attribution error gap between normal and anomalous data, making it particularly suitable for this dataset. As for AUC results, MADLLM outperforms all the baseline methods on all five datasets. MADLLM can improve the F1 score by 1.27\% and improve the AUC result by 2.55\% on average compared with the best performance of the baseline models over the five datasets.

Since models usually have to face unknown pattern anomalies, we also test the few-shot learning (learning with limited training data) ability of MADLLM. Specifically, for the five MTS datasets above, we use 20\% of the training datasets to train the models. The validation and test datasets stay the same.
The results are shown in Table \ref{table:few shot}. Table \ref{table:few shot} illustrates that on each dataset, MADLLM achieves the best F1 score and AUC results. MADLLM has an F1 score 1.26\% higher and AUC 2.65\% higher on average than the best performance of the baseline models over all the datasets with 20\% training data. The results prove the ability of MADLLM to face unknown pattern anomalies. Compared to Table \ref{table:performance}, the LLM-based methods (SFT, AnomalyLLM, LLMAD, and MADLLM) exhibit smaller decreases in F1 score and AUC compared with other methods. This is because LLM has strong generalization capabilities and can achieve good performance even with limited training data.

We also compare the training time of MADLLM with baseline methods. The experiment is conducted on the SMD dataset. As is shown in Table \ref{tab:training time}, compared with all the methods other than SFT, the training time of MADLLM is 95.66\%-99.46\% lower. The reason why SFT and MALLM can achieve such low training time is that they only require fine-tuning of certain parts of the pre-trained LLM. Since MADLLM also involves fine-tuning the contrastive learning encoder, its training time is slightly higher than SFT. However, Table \ref{table:performance} shows that MADLLM increases SFT by 9.2\% in F1 score and 7.8\% in AUC on average over all the datasets. Therefore, compared with SFT, MADLLM achieves a \textbf{significant improvement in accuracy} at the cost of a \textbf{minimal increase in training time}.

\subsection{Ablation Study}
\begin{figure}[t]
    \centering
    \subfloat[F1 score results]{
    \label{fig:ablation_F1}
    \begin{minipage}[b]{0.24\textwidth}
            \includegraphics[width=1\textwidth]{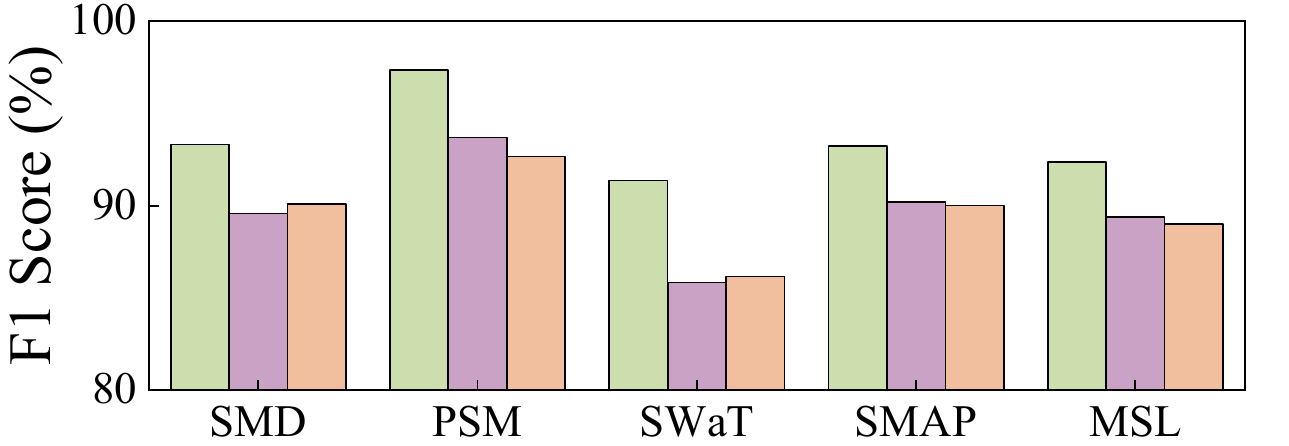}    
        \end{minipage} 
    }
    \subfloat[AUC results]{
    \label{fig:ablation_AUC}
    \begin{minipage}[b]{0.24\textwidth}
            \includegraphics[width=1\textwidth]{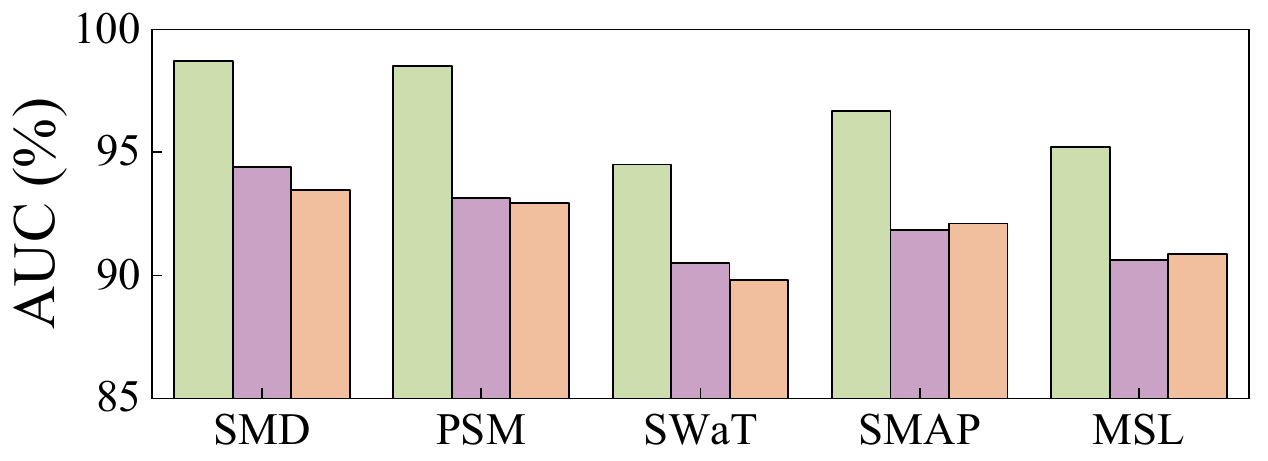}    
        \end{minipage} 
    }
    
    \caption{Ablation study results on different datasets.}
    \label{fig: ablation}
    \vspace{-0.6cm}
\end{figure}

We further measure the impact of skip embedding and feature embedding in MADLLM, respectively. As is shown in Figure \ref{fig: ablation}, removing feature embedding leads to an average decrease of 3.91\% in F1 score and 3.87\% in AUC, removing skip embedding leads to an average decrease of 4.63\% in F1 score and 4.90\% in AUC. These results prove the effect of the two optimizations.

\subsection{Analysis}

\begin{figure}[t]
    \centering
    \subfloat[]{
    \label{fig: analysis3}
    \begin{minipage}[b]{0.24\textwidth}
            \includegraphics[width=1\textwidth]{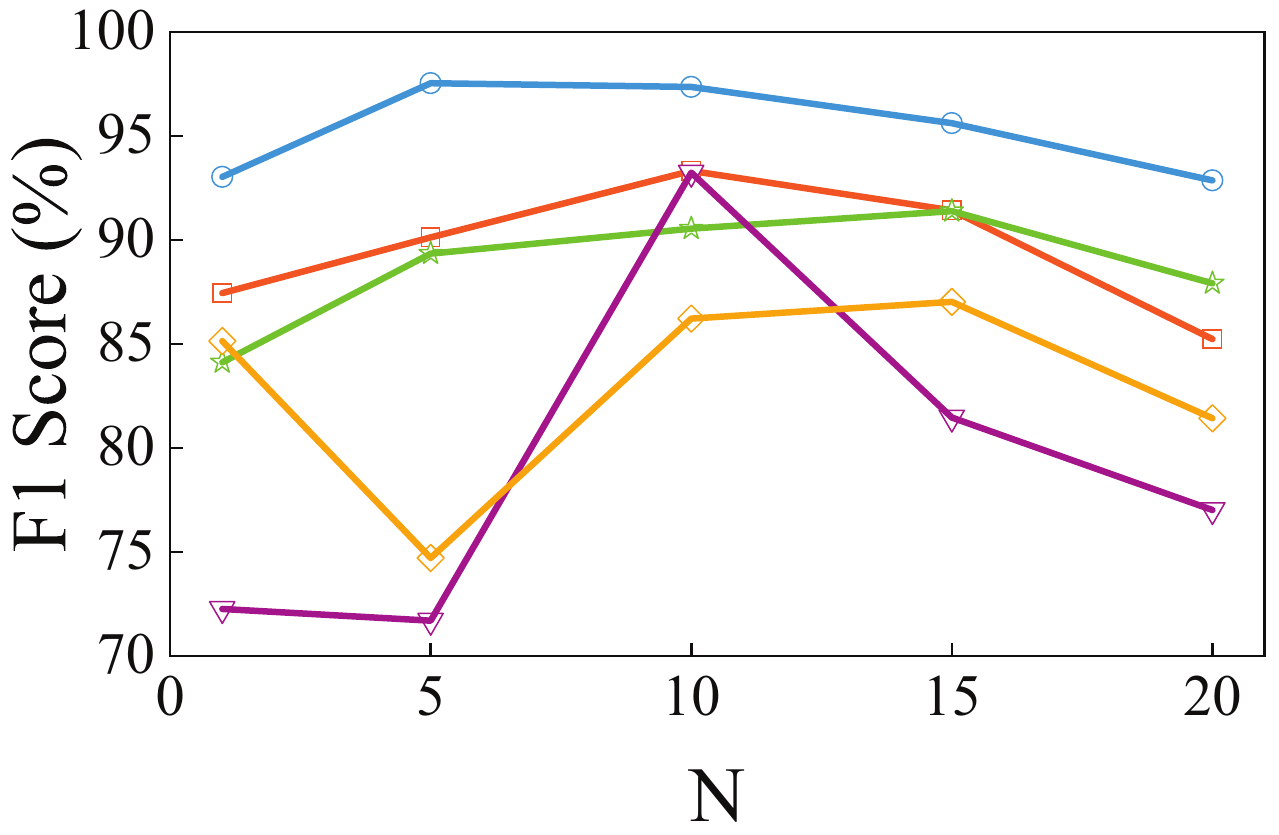}    
        \end{minipage} 
    }
    \subfloat[]{
    \label{fig: analysis6}
    \begin{minipage}[b]{0.24\textwidth}
            \includegraphics[width=1\textwidth]{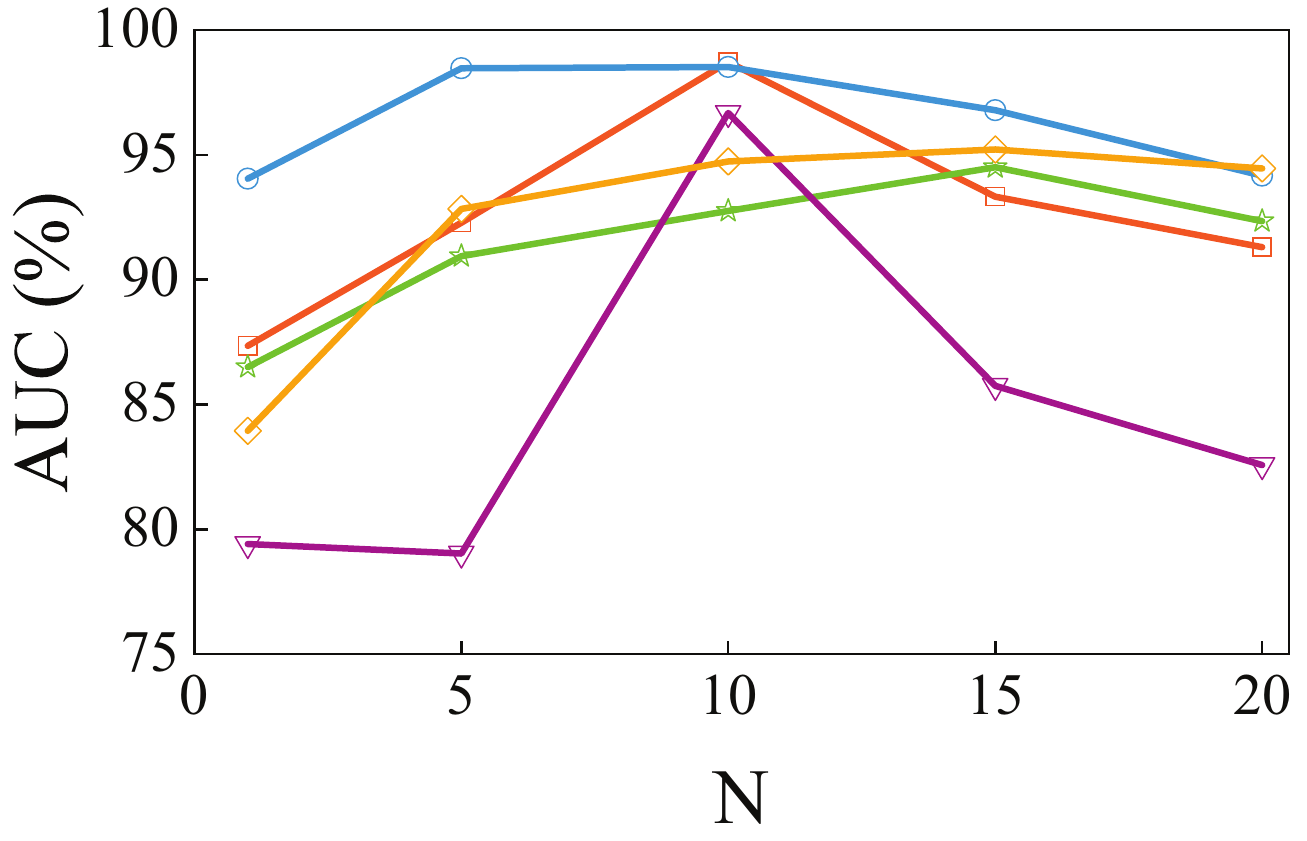}    
        \end{minipage} 
    }
    \caption{(a) The impact of hyperparameter $N$ on F1 score. (b) The impact of hyperparameter $N$ on AUC.}
    \label{fig: analysis}
    \vspace{-7mm}
\end{figure}

We test the F1 score and AUC of MADLLM under different numbers of negative patches selected in the feature embedding triplet loss, i.e., the hyperparameter $N$ in Section \ref{subsec: feature embedding}. The results are shown in Figure \ref{fig: analysis}. When $N$ is small, the F1 scores and AUC results are also quite low. This is because the number of negative patches is too small, and feature embedding cannot yet entirely exert its effect. As $N$ increases, both the F1 score and AUC gradually increase as well. However, when $N$ is too large, the F1 score and AUC results of MADLLM decrease due to overfitting. 

\begin{figure}[t]
    \centering
    \includegraphics[width=0.48\textwidth]{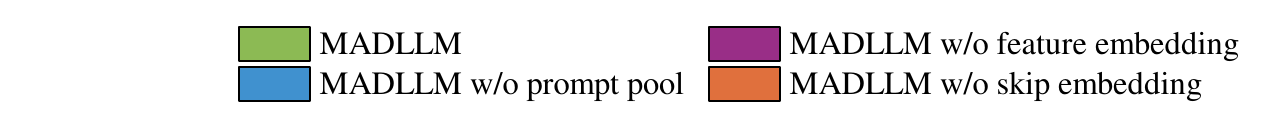}
    \label{fig:enter-label}
    \vspace{-0.9cm}
\end{figure}
\begin{figure}[t]
    \centering
    \includegraphics[width=0.48\textwidth]{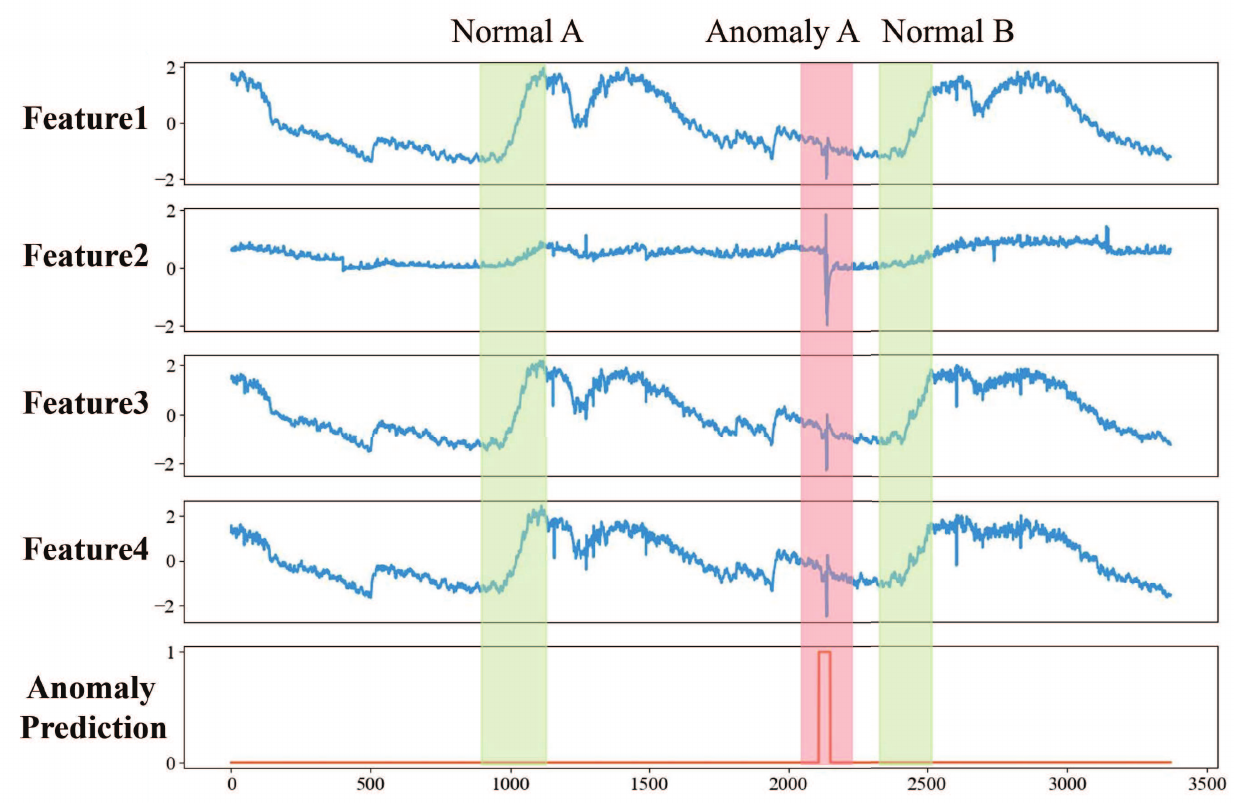}
    \caption{Case study of MADLLM on the SMD dataset.}
    \label{fig: case study}
    \vspace{-0.7cm}
\end{figure}

To make a more concrete evaluation of MADLLM, we provide
a case study on how MADLLM detects anomaly periods for
given real-world MTS datasets. This case study is conducted
using a segmentation of the SMD dataset, similar to the experiment in Figure \ref{fig:challenge}. As is shown in Figure \ref{fig: case study}, MADLLM can accurately detect the true anomaly (Anomaly A). Note that in Normal A and Normal B, \textbf{Feature1, 3, and 4} all show an obvious spike, but \textbf{Feature2} does not. Such a situation is because of a periodically coming task, but these two time periods can easily be misidentified as anomalous. Our MADLLM can effectively prevent such misjudgments.

\section{Conclusion}
\label{sec: conclusion}
In this paper, we introduce MADLLM, a novel multivariate anomaly detection method based on pre-trained LLMs. We implement two key optimizations: ({\romannumeral1}) Skip Embedding: we rearrange the patch processing order in traditional patch embedding method to avoid the pre-trained LLMs forgetting the information of features input in the distant past. ({\romannumeral2}) Feature Embedding: We employ a contrastive learning encoder to learn the correlations among different features, where we design a triplet loss to update the encoder. Experimental results demonstrate that our method outperforms state-of-the-art anomaly detection methods in various public datasets.

\section{Acknowledgements}
This work was sponsored by the Key Research and Development Program of Guangdong Province under grant No. 2021B0101400003, the National Key Research and Development Program of China under Grant No.2023YFB4502701, the China Postdoctoral Science Foundation under Grant No.2024M751011, the Postdoctor Project of Hubei Province under Grant No.2004HBBHCXA027.

\begin{figure}[t]
    \centering
    \includegraphics[width=0.4\textwidth]
    {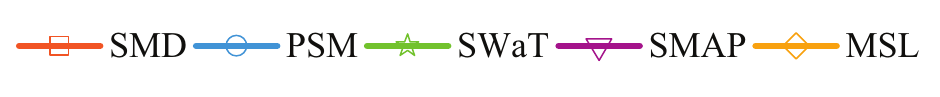}
    \vspace{-1cm}
    \label{fig:enter-label}
\end{figure}

\bibliographystyle{IEEEbib}
\bibliography{icme2025references}

\end{document}